\pdfoutput=1

\documentclass[11pt]{article}

\usepackage{acl}

\usepackage{times}
\usepackage{latexsym}

\usepackage[T1]{fontenc}
\usepackage{CJKutf8}
\usepackage{multirow}
\usepackage[export]{adjustbox}

\usepackage[utf8]{inputenc}

\usepackage{microtype}
\usepackage{natbib}
\usepackage{subcaption}

\title{K-MHaS: A Multi-label Hate Speech Detection Dataset \\in Korean Online News Comment}

\author{
    Jean Lee\textsuperscript{1} \quad
    Taejun Lim\textsuperscript{1} \quad
    Heejun Lee\textsuperscript{3} \quad
    Bogeun Jo\textsuperscript{3} \quad \\
    \textbf{Yangsok Kim\textsuperscript{4}} \quad
    \textbf{Heegeun Yoon\textsuperscript{5}} \quad
    \textbf{Soyeon Caren Han\textsuperscript{1,2}}\thanks{\, Corresponding author (caren.han@sydney.edu.au)} \\
    The University of Sydney\textsuperscript{1} \quad The University of Western Australia\textsuperscript{2} \\
    BigWave AI\textsuperscript{3} \quad Keimyung University\textsuperscript{4} \quad National Information Society Agency\textsuperscript{5} \\
    \texttt{\{jean.lee, tlim6535, caren.han\}@sydney.edu.au} \\
    \texttt{\{leej3471, iambgjo\}@gmail.com} \\
    \texttt{yangsok.kim@kmu.ac.kr} \quad
    \texttt{hgyoon@nia.or.kr}
}

\begin{document}
\maketitle
\begin{abstract}
Online hate speech detection has become an important issue due to the growth of online content, but resources in languages other than English are extremely limited. We introduce K-MHaS\footnote{The dataset is available at https://github.com/adlnlp/K-MHaS.}, a new multi-label dataset for hate speech detection that effectively handles Korean language patterns. The dataset consists of 109k utterances from news comments and provides a multi-label classification using 1 to 4 labels, and handles subjectivity and intersectionality. We evaluate strong baselines on K-MHaS. KR-BERT with a sub-character tokenizer outperforms others, recognizing decomposed characters in each hate speech class.
\end{abstract}

\section{Introduction}
The growth of online content including social media \citep{zampieri2020semeval}, news comments \citep{GaoH17}, Wikipedia \citep{wulczyn2017ex}, and in-game chat \citep{weld2021conda} presents challenges in detecting hate speech using advanced Natural Language Processing. Hate speech is language that attacks or diminishes individuals or groups based on certain characteristics such as physical appearance, religion, gender, or other attributes, and it can occur across different linguistic styles \citep{fortuna2018survey}. Hate speech detection is intrinsically a complex task \citep{wang2020detect} due to the fuzzy boundary with other overlapping concepts such as abusive language \citep{NobataTTMC16}, toxic comments \citep{wulczyn2017ex}, or offensive language \citep{DavidsonWMW17}.

Recently, the rise in popularity of Korean TV, movies, and music (e.g. Squid Game, BTS) has led to many young people showing an interest in learning Korean. This phenomenon could result in exposure to harmful content and hate speech in Korean. 
However, (1) the most common language in hate speech research is English and only limited resources are available in other languages such as Arabic \citep{mubarak2017abusive}, Dutch \citep{caselli2021dalc}, and Korean \citep{moon2020beep}. In addition, most datasets are annotated (2) using a single label classification of particular aspects, even though the subjectivity of hate speech cannot be explained with a mutually exclusive annotation scheme.

We propose K-MHaS, a Korean multi-label hate speech detection dataset that allows overlapping labels associated with intersectionality, a concept from sociology that identifies combined attributes \citep{crenshaw1989demarginalizing}. Our dataset consists of 109,692 utterances from Korean online news comments, labeled with 8 fine-grained hate speech classes. K-MHaS is compatible with previous work on hate speech in other languages, by providing binary classification and multi-label classification from 1(one) to 4(four) labels.



We investigate the K-MHaS dataset by analyzing label distribution, keywords, and label pairs. In addition, we provide strong baseline pre-trained language models using Multilingual-BERT, KoELECTRA, KoBERT, and KR-BERT, and compare the results using six metrics for multi-label classification tasks. Overall, the KoELECTRA model achieves the best performance for all labels, indicating the effects of the pre-training data source. The KR-BERT with a sub-character-level tokenizer outperforms the others on several label pairs, showing that decomposing various Korean characters is essential for the task. Our contribution can be summarized as follows:

\begin{table*}[t]
\centering
\small
\setlength{\tabcolsep}{1.2mm}{
\renewcommand{\arraystretch}{1.1}
\begin{tabular}{lllclc}
\hline
\textbf{Publication}& 
\textbf{Language}& 
\textbf{Source}& 
\textbf{Data size}& 
\textbf{Labels}& 
\textbf{M-label} \\ 
\hline
\citet{WaseemH16}
& English
& Twitter
& 16.2k
& Sexism, Racism, Neither
& N
\\ \hline
\citet{DavidsonWMW17}
& English
& Twitter
& 24.8k
& Hate Speech, Offensive, Neither
& N
\\ \hline
\citet{wulczyn2017ex}
& English
& \begin{tabular}[c]{@{}l@{}}Wikipedia\\ comments\end{tabular}
& 115k 
& \begin{tabular}[c]{@{}l@{}}Toxic, Severe Toxic, Obscene, Threat,  \\ Insult, Identity Hate, Neutral\end{tabular} 
& Y
\\ \hline
\citet{ibrohim2019multi}
& Indonesian 
& Twitter
& 11k 
& \begin{tabular}[c]{@{}l@{}l@{}}(a)  Individual, Group\\ (b) Religion, Race, Pysical, Gender, Other\\(c) Weak, Moderate, Strong Hate Speech\end{tabular} 
& P
\\ \hline
\citet{fortuna2019hierarchically}
& Portuguese 
& Twitter
& 5.6k
& \begin{tabular}[c]{@{}l@{}l@{}}(a) Hate Speech, Not Hate Speech\\ (b) Sexism, Body, Origin, Homophobia, Racism,\\ Ideology, Religion, Health, Other-Lifestyle \end{tabular} 
& P
\\ \hline
\citet{ousidhoum2019multilingual} 
& \begin{tabular}[c]{@{}l@{}l@{}}English\\ French\\ Arabic \end{tabular} 
& Twitter
& \begin{tabular}[c]{@{}l@{}l@{}}6k (EN)\\4k (FR)\\3k (AR) \end{tabular}
& \begin{tabular}[c]{@{}l@{}l@{}}Labels for five different aspects\\ (a) Directness, (b) Hostility, (c) Target, \\ (d) Group, and (e) Annotator\end{tabular}
& P 
\\ \hline
\citet{moon2020beep}
& Korean 
& \begin{tabular}[c]{@{}l@{}}News \\ comments\end{tabular}  
& 9k
& \begin{tabular}[c]{@{}l@{}}(a) Hate Speech, Offensive, None\\ (b) Gender, Others, None\end{tabular}
& N
\\ \hline
\textbf{Ours}
& \textbf{Korean} 
& \begin{tabular}[c]{@{}l@{}}\textbf{News} \\ \textbf{comments}\end{tabular}  
& \textbf{109k}
& \textbf{\begin{tabular}[c]{@{}l@{}l@{}}(a) Hate Speech, Not Hate Speech \\(b) Politics, Origin, Physical, Age, Gender \\ Religion, Race, Profanity, Not Hate Speech\end{tabular}}
& \textbf{Y}
\\ \hline
\end{tabular}
}
\setlength{\belowcaptionskip}{-10pt}
\caption{Comparison of datasets. A "M-label" indicates a multi-label annotation scheme that allows overlapping labels for intersectionality (P = partially applied). The (a) - (e) indicates a layer containing a single label from each aspect.}
\label{tab:dataset_comparisons}
\end{table*}

\begin{itemize}
\item We propose a large size Korean multi-label hate speech detection dataset that represents Korean language patterns effectively;
\item We propose a multi-label hate speech annotation scheme, which can handle the subjectivity of hate speech and the intersectionality;
\item We evaluate strong baseline experiments on our dataset using Korean-BERT-based language models with six different metrics.
\end{itemize}

\section{Korean Multi-label Hate Speech Detection Dataset (K-MHaS)}\label{data_kmhas}

Our dataset is based on the Korean online news comments available on Kaggle \footnote{https://www.kaggle.com/datasets/junbumlee/kcbert-pretraining-corpus-korean-news-comments} and Github \footnote{https://github.com/kocohub/korean-hate-speech}. The unlabeled raw data was collected between January 2018 and June 2020. In order to curate the data, we randomly select more than 109,692 news comments. Our data preprocessing is designed to tokenize a Korean character and to filter the length. We remove URLs and bad characters (e.g. U+1100 to U+11FF - Hangul Jamo) using regular expressions while keeping uppercase and lowercase letters in English and emoji. We discard sentences with fewer than 10 characters as it is often only one word. For the data derived from online comments, we normalized repeated characters by truncating their number of consecutive repetitions to two.

\paragraph{Multi-label Annotation}
We consider a multi-label annotation scheme in order to deliver fine-grained hate speech categories and intersectionality from the overlapping labels. The annotation scheme has two layers: (a) binary classification (\textit{`Hate Speech' or `Not Hate Speech'}) and (b) fine-grained classification (\textit{8 labels or `Not Hate Speech'}). For the fine-grained classification, a `Hate Speech' class from the binary classification is broken down into 8 classes associated with the hate speech category
\footnote{
\begin{CJK}{UTF8}{mj}
Fine-grained labels (matching in Korean) : Politics (정치성향차별), Origin (출신차별), Physical (외모차별), Age (연령차별), Gender (성차별), Religion (종교차별), Race (인종차별), and Profanity (혐오욕설)
\end{CJK}}. 
As shown in Table \ref{tab:dataset_comparisons}, this scheme allows non-exclusive concepts, accounting for the overlapping shades of given categories. We select the 8 hate speech classes in order to reflect the social and historical context as the nature of hate speech is different in each language \citep{kang2020hate}. For example, the ‘\textit{politics}’ class is chosen due to a significant influence on the style of Korean hate speech.

\paragraph{Annotation Instructions}
Given the subjectivity of the task and our annotation scheme, we perform a preliminary round to identify the topics of hate speech and develop annotation instructions. We begin with the common categories of hate speech found in literature and match the keywords for each category. After the preliminary round, we investigate the results to merge or remove labels in order to provide the most representative subtype labels of hate speech contextual to the cultural background. Our annotation instruction includes the criteria as follows: \textbf{Politics}: hate speech based on political stance; \textbf{Origin}: hate speech based on place of origin or identity; \textbf{Physical}: hate speech based on physical appearance (e.g. body, face) or disability; \textbf{Age}: hate speech based on age; \textbf{Gender}: hate speech based on gender or sexual orientation (e.g. woman, homosexual); \textbf{Religion}: hate speech based on religion; \textbf{Race}: hate speech based on ethnicity; \textbf{Profanity}: hate speech in the form of swearing, cursing, cussing, obscene words, or expletives; or an unspecified hate speech category from above; and \textbf{Not Hate Speech}.


Our annotation instructions explain a two-layered annotation to (a) distinguish hate and not hate speech, and (b) the categories of hate speech. Annotators are requested to consider given keywords or alternatives of each category within social, cultural, and historical circumstances. For example, a comment using the word \textit{“women”} is not hate speech, whereas, if it is critical of \textit{“women”} or uses language that attacks the group, it is classified as \textit{‘gender’}. Notably, we annotate multi-labels if a comment includes several hate speech categories. Since hate speech can be varied, any comments in the form of swearing or cursing are marked as \textit{‘profanity’}. For instance, a comment containing hate speech about appearance, political stance, and gender in profane language (e.g. \textit{"fuck you ugly communist bitch."})\footnote{
\begin{CJK}{UTF8}{mj}
(Korean) “면상도 개 조가치 생겼네 개빨갱이년”
\end{CJK}} is labeled within \textit{‘physical’, ‘politics’, ‘gender’} and \textit{‘profanity’} classes.

\begin{figure}[t]
  \centering
  \includegraphics[width=\linewidth]{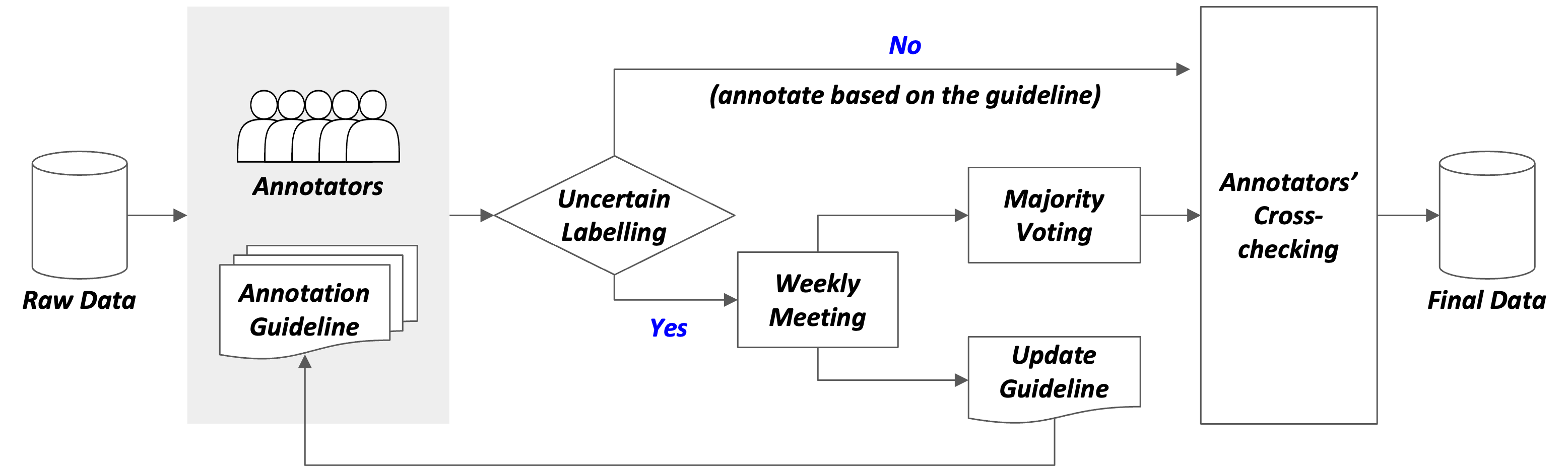}
  \caption{Overview of Annotation Process.}
  \label{fig:anno_process}
\end{figure}

\paragraph{Annotation Process}
Five native speakers were recruited for manual annotation in both the preliminary and main rounds. During the preliminary round, we facilitated the annotation instructions by conducting an annotators’ discussion and providing some examples of keywords for each class. As shown in Figure \ref{fig:anno_process}, we introduced an iterative process that enables faster annotation in the main round. We provided an ‘\textit{uncertain}’ additional field that was used for the unspecified label in annotation guidelines or when the annotator had difficulties in choosing labels. Any ‘\textit{uncertain}’ labeled data was flagged by individual annotators, then reviewed by five annotators. The final labels were chosen based on the majority vote, and the annotation guidelines were updated to handle similar cases. Additionally, the other labeled data was reviewed, in line with the annotation guideline by two random annotators for the final dataset. The inter-annotator agreement returns an average Cohen Kappa score of 0.892, indicating substantial agreement \citep{siegel1988nonparametric}.


\section{Dataset Analysis}
K-MHaS dataset contains 109,692 comments as shown in Table \ref{tab:data_stat}. For binary classification, the proportion of the ‘\textit{hate speech}’ (45.7\%) and ‘\textit{not hate speech}’ (54.3\%) satisfies data balancing. The ‘\textit{hate speech}’ label consists of a single label (33.2\%) and multi-labels (12.4\%), containing from 2 to 4 labels. Other hate speech datasets reviewed have an approximate ratio of ‘\textit{hate speech}’ to ‘\textit{not hate speech}’ of around 40\% \citep{vidgen2020directions}. Our dataset is consistent with this figure, where the ‘\textit{hate speech}’ in a single label to ‘\textit{not hate speech}’ ratio is 38\%.

\begin{table}[t]
\centering
\scriptsize
\setlength{\tabcolsep}{4mm}{
\renewcommand{\arraystretch}{1.2}
\begin{tabular}{ll|r}
\hline
\multicolumn{2}{l|}{\textbf{Label Types}} &\textbf{Count (\%)} \\ \hline
\multicolumn{2}{l|}{Total Utterances} & \textbf{109,692 (100\%)}     \\ \hline
\multicolumn{1}{l}{\multirow{4}{*}{\begin{tabular}[c]{@{}l@{}}Multi-label \\ (Hate Speech)\end{tabular}}} 
& 1 label (Single)    & 36,470 (33.2\%)     \\ \cline{2-3} 
\multicolumn{1}{l}{} & 2 labels    & 12,073 (11.0\%)       \\ \cline{2-3} 
\multicolumn{1}{l}{} & 3 labels    & 1,440 (1.3\%)       \\ \cline{2-3} 
\multicolumn{1}{l}{} & 4 labels    & 94 (0.1\%)          \\ \hline
\multicolumn{2}{l|}{Not Hate Speech}& 59,615 (54.3\%)     \\ \hline

\end{tabular}
}
\caption{Dataset Statistics. The total is the combination of all ‘\textit{hate speech}’ and ‘\textit{not hate speech}’ label. Together the ‘\textit{hate speech}’ label makes up 45.7\% of the data.}
\label{tab:data_stat}
\end{table}

\begin{table}[t]
\centering
\scriptsize
\setlength{\tabcolsep}{4mm}{
\renewcommand{\arraystretch}{1.2}
\begin{tabular}{l|rr}
\hline
\textbf{Class} & \textbf{Count - Single (\%)} & \textbf{Count - Multi (\%)} \\ \hline
Politics       & 6,931 (19.0\%)               & 4,961 (17.2\%)              \\ 
Origin          & 5,739 (15.7\%)               & 4,458 (15.5\%)              \\ 
Physical       & 5,443 (14.9\%)               & 3,364 (11.7\%)              \\ 
Age            & 4,192 (11.5\%)               & 3,178 (11.0\%)              \\ 
Gender         & 3,348 (9.2\%)                & 4,696 (16.3\%)              \\ 
Religion       & 1,862 (5.1\%)                & 513 (1.8\%)                 \\ 
Race           & 160 (0.4\%)                  & 163 (0.6\%)                 \\ 
Profanity      & 8,795 (24.1\%)               & 7,509 (26.0\%)              \\ 
\hline
\end{tabular}
}
\caption{Fine-grained label distributions on \textit{hate speech} labels. A ‘\textit{not hate speech}’ label is not included. A single means 1 label and a multi is the sum of 2, 3, and 4 labels. A multi-labeled data counts each overlapping class.}
\setlength{\belowcaptionskip}{-15pt}
\label{tab:label_dist}
\end{table}





\paragraph{Label Distribution}
Table \ref{tab:label_dist} shows the fine-grained label distribution across our K-MHaS. For both single (s) and multi-label (m) distribution, the ‘\textit{profanity}’ class (24.1\%-s, 26.0\%-m) is more frequent than any other class, indicating that swear words are critical for detecting hate speech. Also, the ‘\textit{religion}’ (5.1\%-s, 1.8\%-m) and ‘\textit{race}’ (0.4\%-s, 0.6\%-m) classes are the smallest portions in both distributions, which are significantly more common in other hate speech datasets. This difference could be because Korea is a highly homogenous monoculture with little variation in race and religion \citep{kang2020hate}. Interestingly, the ‘\textit{gender}’ class (16.3\%) occurs at almost twice the frequency in a multi-label distribution, compared to a single label distribution (9.2\%). This indicates that gender-based hate speech is used extensively in combined aspects.

\paragraph{Keyword Analysis}
\begin{CJK}{UTF8}{mj}
To understand the lexical aspects, we list the top 5 keywords for each hate speech category in Table \ref{tab:keywords}, identifying which tokens are highly associated with each class. In the ‘\textit{politics}’ class, we find that far-right extremism is dominant, and new tokens such as \textit{“catastrophe”} [jae ang](재앙) appears related to the former president's given name ([jae in]) as the two words are near-homophones. Across all classes, one-word tokens are often used in their stem form to modify the meanings of other words. For example, a token [teul] (틀) comes from the word \textit{“denture”} [teulni] (틀니) which is used as an offensive reference to the elderly. In addition, one-word tokens can be used as a prefix (e.g. “dog” [gae] (개)) or a suffix (e.g. “insect” [chung] (충)), and combined with other neutral words to create a new offensive term.
\end{CJK}

\begin{table}[t]
\centering
\scriptsize
\setlength{\tabcolsep}{2mm}{
\renewcommand{\arraystretch}{1.2}
\begin{CJK}{UTF8}{mj}
\begin{tabular}{c|cccc}
\hline
\textbf{Rank} & \textbf{Politics} & \textbf{Origin}   & \textbf{Physical} & \textbf{Age}       \\ \hline
1             & 재앙 (1427)         & 짱깨 (615)          & 얼굴 (962)          & 틀 (1918)           \\ 
2             & 문재인 (951)         & 전라도 (596)         & 돼지 (772)          & 나이 (599)           \\ 
3             & 좌파 (464)          & 중국 (539)          & 여자 (294)          & 노인 (139)           \\ 
4             & 좌빨 (402)          & 쪽 (448)           & 성형 (216)          & 충 (112)            \\ 
5             & 빨갱이 (367)         & 짱 (446)           & 관상 (183)          & 놈 (106)            \\ \hline
\textbf{Rank} & \textbf{Gender}   & \textbf{Religion} & \textbf{Race}     & \textbf{Profanity} \\ \hline
1             & 여자 (1704)         & 개독 (526)          & 흑인 (44)           & 새끼 (1103)          \\ 
2             & 남자 (990)          & 신천지 (460)         & 백인 (32)           & 년 (1014)           \\ 
3             & 페미 (172)          & 사이비 (409)         & 양키 (32)           & 지랄 (564)           \\ 
4             & 맘충 (138)          & 종교 (305)          & 깜둥이 (19)          & 개 (459)            \\ 
5             & 여성 (134)          & 예수 (227)          & 놈 (13)            & 놈 (404)            \\ \hline
\end{tabular}
\end{CJK}
}
\setlength{\belowcaptionskip}{-10pt}
\caption{Top 5 keywords associated with each fine-grained label. The number in brackets is the token count. The keyword analysis is from the total dataset and different from some examples in annotation guidelines.}
\label{tab:keywords}
\end{table}


\begin{figure}[t]
\centering
\begin{subfigure}{.33\linewidth}
  \centering
  \includegraphics[width=\linewidth]{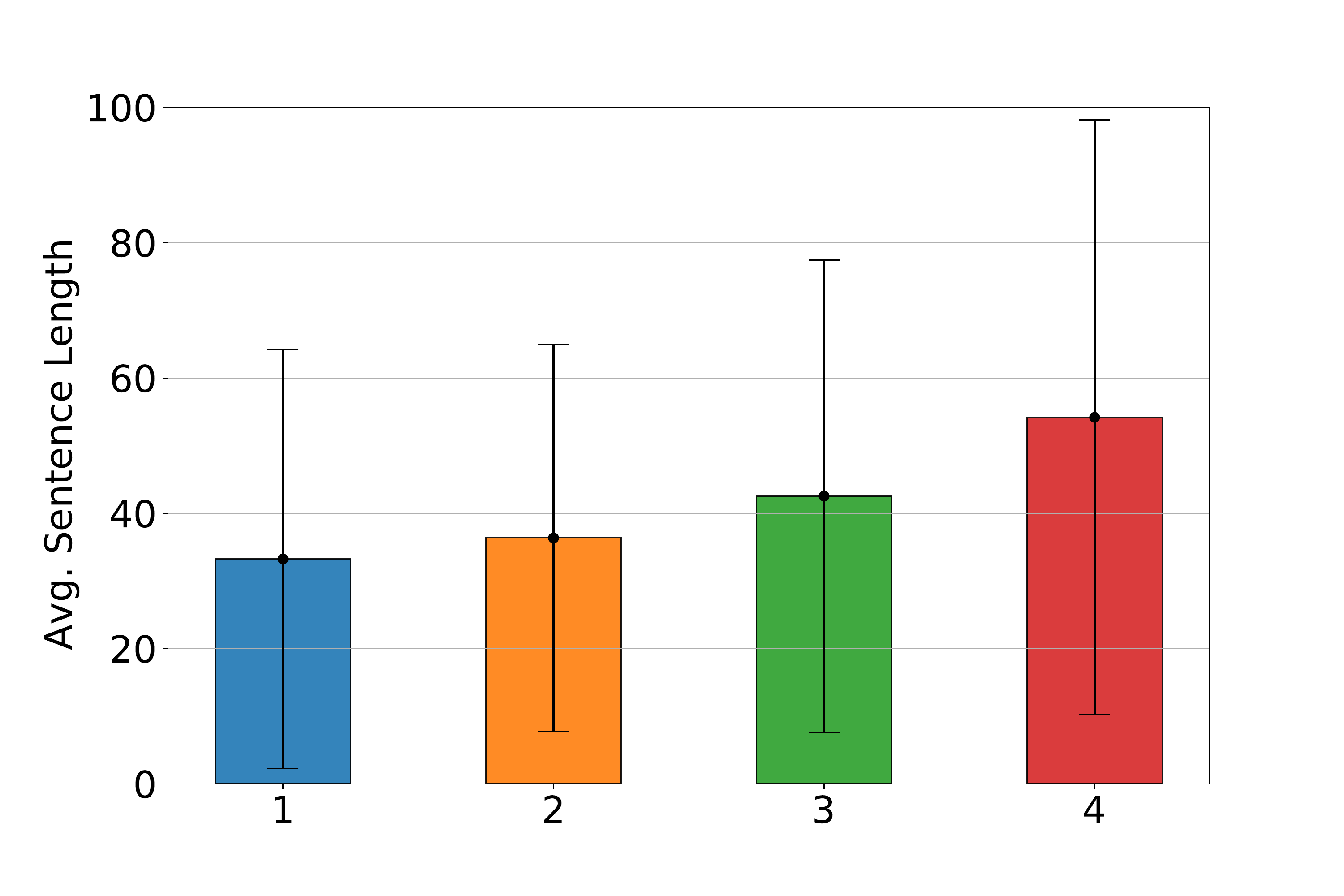}
  \caption{Label types}
  \label{fig:label_avg_length}
\end{subfigure}%
\begin{subfigure}{.33\linewidth}
  \centering
  \includegraphics[width=\linewidth]{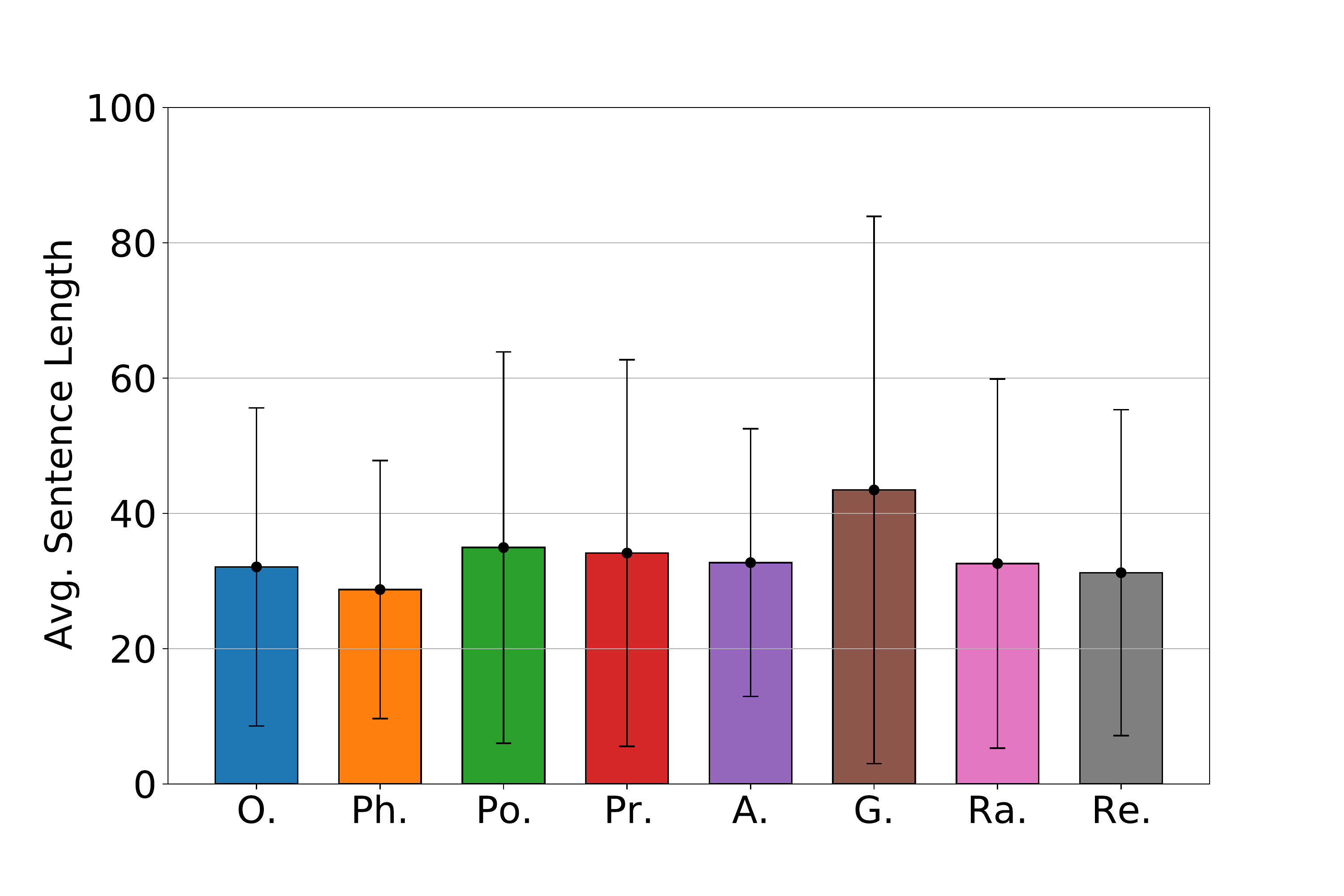}
  \caption{Class types I}
  \label{fig:class_single_label_avg_length}
\end{subfigure}%
\begin{subfigure}{.33\linewidth}
  \centering
  \includegraphics[width=\linewidth]{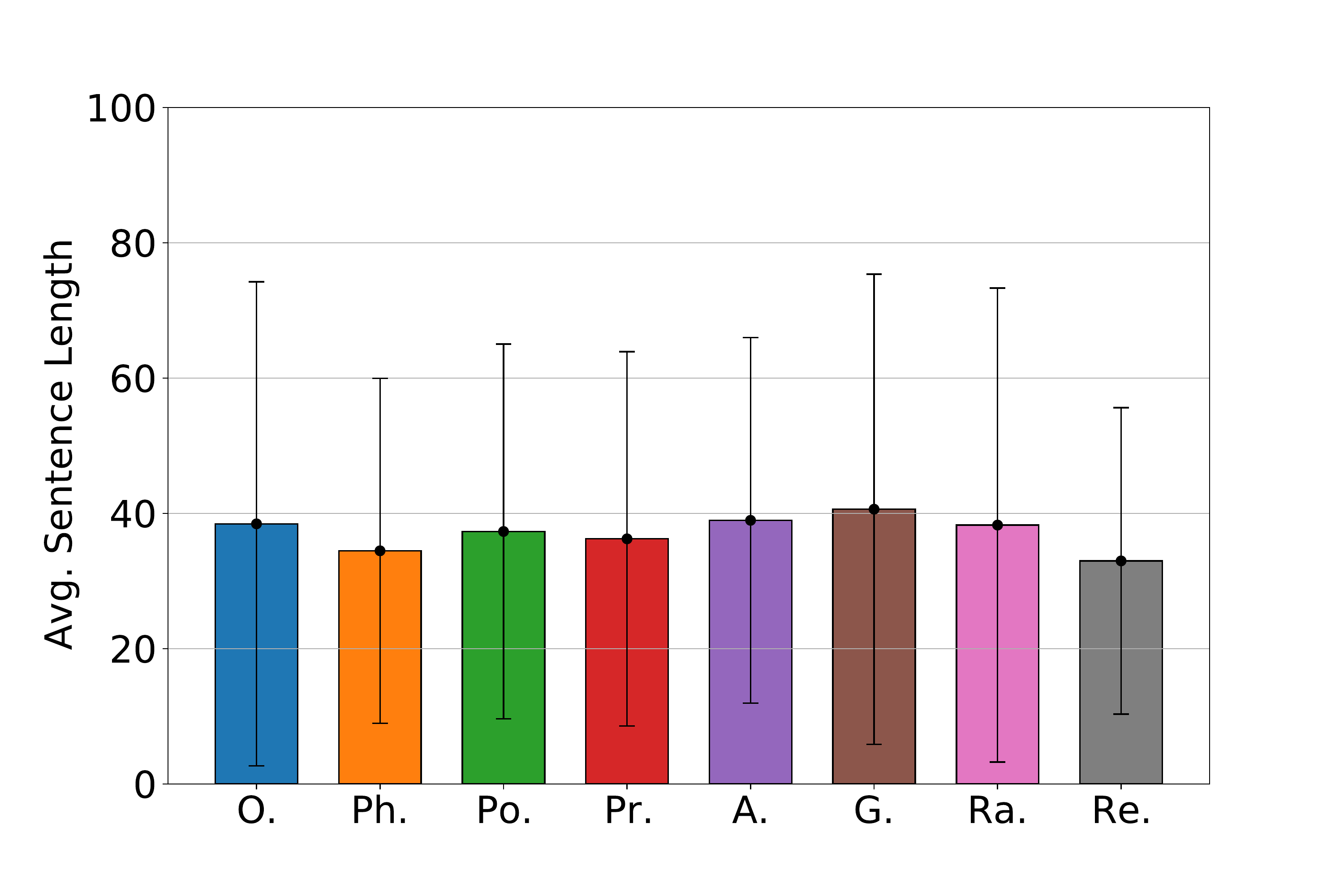}
  \caption{Class types II}
  \label{fig:class_multiple_label_avg_length}
\end{subfigure}
\setlength{\belowcaptionskip}{-10pt}
\caption{Average utterance length. (a) label types from 1 to 4 labels. 8 class types (b) in a single label and (c) in multi-labels.}
\label{fig:average_sentence_length}
\end{figure}

\paragraph{Label Pair Analysis}
Figure \ref{fig:average_sentence_length} shows the average length of utterance by label count and class type. The total average length of an utterance is 33 tokens. An increase in the number of labels shows an increasing trend in utterance length, indicating that multi-labeled hate speech contains more linguistic content. The \textit{‘gender’} class has relatively longer lengths (43 tokens) compared to other classes in a single label, whereas all multi-labels utterances have a similar length. This indicates that the gender class has different linguistic features.

\begin{table}[t]
\centering
\resizebox{\linewidth}{!}{
\renewcommand{\arraystretch}{1.2}
\begin{tabular}{l|cccccc}
\hline
\textbf{Model}      
& \textbf{F1 (macro)}
& \textbf{F1 (micro)} 
& \textbf{F1 (weighted)} 
& \textbf{E.M.} 
& \textbf{AUC} 
& \textbf{H.L. (↓)} \\ \hline
\textbf{BERT} & 0.6912 & 0.8139 & 0.8119 & 0.7579 & 0.8878 & 0.0464 \\ 
\textbf{KoELECTRA} & 0.7245 & \underline{0.8493} & \textbf{0.8480} & \textbf{0.7994} & \textbf{0.9122} & \underline{0.0380} \\ 
\textbf{KoBERT} & \textbf{0.7651} & 0.8413 & 0.8424 & \underline{0.7926} & \underline{0.9083} & 0.0401 \\ 
\textbf{KR-BERT-c} & \underline{0.7444} & \textbf{0.8500} & \underline{0.8470} & 0.7901 & 0.9028 & \textbf{0.0368} \\ 
\textbf{KR-BERT-s} & 0.7245 & 0.8445 & 0.8437 & 0.7825 & 0.9076 & 0.0390 \\ \hline
\end{tabular}
}
\caption{Overall multi-label classification performance on K-MHaS for the five baseline models at epoch 4 (E.M.:Exact Match, H.L.:Hamming Loss / KR-BERT-*: c = character-level, s = sub-character-level)}
\label{tab:overall_ev}
\end{table}

\begin{table}[t]
\centering
\resizebox{\linewidth}{!}{
\renewcommand{\arraystretch}{1.2}
\begin{tabular}{c|c|ccccc}
\hline
\textbf{Metric} & \textbf{\# labels} & \textbf{BERT} & \textbf{KoELECTRA} & \textbf{KoBERT} & \textbf{KR-BERT-c} & \textbf{KR-BERT-s} \\ \hline
\multirow{4}{*}{\textbf{\begin{tabular}[c]{@{}c@{}}F1\\ (micro)\end{tabular}}} & 1 & 0.8190 & \underline{0.8490} & 0.8320 & \textbf{0.8553} & 0.8392 \\
 & 2 & 0.8043 & 0.8612 & \textbf{0.8854} & 0.8405 & \underline{0.8703} \\
 & 3 & 0.7517 & 0.7987 & \underline{0.8290} & 0.7827 & \textbf{0.8329} \\
 & 4 & 0.7093 & 0.7044 & 0.6832 & \underline{0.7439} & \textbf{0.7771} \\ \hline
\end{tabular}}
\setlength{\belowcaptionskip}{-10pt}
\caption{A breakdown of F1 for multi-label classification from 1 to 4 labels.}
\label{tab:breakdown_labels}
\end{table}

\section{Experiment Setup}
\label{sec:experiment}
\paragraph{Data Preparation} We split the data into train/test in the proportions of 0.8/0.2. From the training set, we randomly select 0.1  as a validation set (78,977/8,776/21,939 samples for train/val/test sets, preserving the class proportion). The data passed to the models is the preprocessed sentences and binary label vectors.


\begin{table*}[t]
\centering
\resizebox{\textwidth}{!}{
\renewcommand{\arraystretch}{1.2}
\begin{tabular}{l|c|ccccc|ccccc}
\hline
\multicolumn{1}{l|}{\textbf{Label Pairs}} & \textbf{\# pairs} & \multicolumn{5}{c|}{\textbf{F1 (macro)}} & \multicolumn{5}{c}{\textbf{F1 (micro)}} \\
\hline
\multicolumn{2}{l|}{\multirow{2}{*}{\textbf{Overall Performance (F1)}}} & \textbf{BERT} & \textbf{KoELECTRA} & \textbf{KoBERT} & \textbf{KR-BERT-c} & \textbf{KR-BERT-s} & \textbf{BERT} & \textbf{KoELECTRA} & \textbf{KoBERT} & \textbf{KR-BERT-c} & \textbf{KR-BERT-s} \\ \cline{3-12}
\multicolumn{2}{c|}{} & 0.6912 & 0.7245 & \textbf{0.7651} & \underline{0.7444} & 0.7245 & 0.8139 & \underline{0.8493} & 0.8413 & \textbf{0.8500} & 0.8445 \\
\hline
\textit{Profanity \& Politics} & 323 & 0.1959 & \underline{0.2045} & \textbf{0.2072} & 0.2013 & 0.2034 & 0.8379 & \underline{0.8853} & \textbf{0.9010} & 0.8687 & 0.8616 \\
\textit{Profanity \& Physical} & 311 & 0.1931 & 0.2061 & \underline{0.2115} & 0.2099 & \textbf{0.2121} & 0.8393 & 0.9096 & 0.9331 & \textbf{0.9369} & \underline{0.9334} \\
\textit{Profanity \& Origin} & 269 & 0.1887 & \underline{0.1989} & 0.1987 & 0.1961 & \textbf{0.2050} & 0.8144 & \underline{0.8731} & 0.8729 & 0.8661 & \textbf{0.9070} \\
\textit{Gender \& Origin} & 242 & 0.2035 & 0.2017 & \underline{0.2134} & 0.1905 & \textbf{0.2141} & 0.8920 & 0.8780 & \underline{0.9440} & 0.8354 & \textbf{0.9494} \\
\textit{Politics \& Origin} & 224 & 0.1962 & 0.1976 & \underline{0.1991} & 0.1872 & \textbf{0.2013} & 0.8666 & 0.8714 & \underline{0.8846} & 0.8295 & \textbf{0.8918} \\
\textit{Age \& Politics} & 222 & 0.1996 & \textbf{0.2114} & \underline{0.2104} & 0.1964 & 0.2014 & 0.8765 & \underline{0.9329} & \textbf{0.9357} & 0.8734 & 0.8878 \\
\textit{Gender \& Profanity} & 181 & 0.1895 & \underline{0.1991} & 0.1957 & 0.1911 & \textbf{0.2054} & 0.8157 & \underline{0.8715} & 0.8542 & 0.8450 & \textbf{0.8994} \\
\textit{Gender \& Physical} & 177 & 0.1160 & 0.1833 & \textbf{0.1958} & 0.1813 & \underline{0.1953} & 0.4562 & 0.7867 & \underline{0.8455} & 0.8045 & \textbf{0.8585} \\
\textit{Age \& Profanity} & 132 & 0.1908 & \underline{0.2102} & \textbf{0.2139} & 0.2063 & 0.2043 & 0.8414 & \underline{0.9240} & \textbf{0.9459} & 0.9105 & 0.9095 \\
\textit{Gender \& Age} & 130 & 0.1738 & \underline{0.1781} & \textbf{0.1903} & 0.1517 & 0.1339 & 0.7277 & \underline{0.7452} & \textbf{0.8159} & 0.6368 & 0.5686 \\ \hline
\end{tabular}}
\setlength{\belowcaptionskip}{-10pt}
\caption{F1 score for the top 10 two-label pairs on the K-MHaS dataset for the five pre-trained language models at epoch 4 (\# total label pairs = 2,439 / KR-BERT-*: c = character-level tokenizer, s = sub-character-level tokenizer).}
\label{tab:breakdown_pair}
\end{table*}

\paragraph{Baselines} We select four baselines. 1) \textbf{Multi-BERT} \citep{BERTMultilingual} is pre-trained on Wikipedia in 104 different languages. We adopted the BERT-Base, uses the WordPiece tokenizer and contains 110M parameters and 119K vocabs. 2) \textbf{KoELECTRA} \citep{KoELECTRA} is pre-trained on 34GB Korean news, Korean Wikipedia, Namuwiki (Korean-based wiki) and Modu (Korean corpus data publicly provided by the Korean government). The KoELECTRA-Small-v3 is used with the WordPiece tokenizer and contains 14M parameters and 35K vocabs. 3) \textbf{KoBERT}~\citep{KoBERT} is pre-trained on 54M words from Korean Wikipedia, used the SentencePiece tokenizer, 92M parameters and 8K vocabs. 4) \textbf{KR-BERT} \citep{KRBERT} is pre-trained on 2.47GB corpus with 233M words from Korean Wikipedia and news. We applied either (1) the character-level tokenizer or (2) the sub-character-level tokenizer\footnote{The KR-BERT tokenization variants can be found as follows: https://github.com/snunlp/KR-BERT\#tokenization}.

\paragraph{Evaluation Metrics} In multi-label classification, the prediction contains a set of labels, which means the prediction can be fully correct, partially correct, or fully incorrect. We propose to use the widely used six metrics~\citep{godbole2004discriminative} for conducting our multi-label classification, including F1-[macro, micro, weighted], Exact Match, AUC and Hamming Loss~\citep{sorower2010literature}.

\section{Results}
\paragraph{Evaluation for All Labels}
The overall performance for all labels is provided in Table \ref{tab:overall_ev}. The F1(micro) range between 0.8139 (Multi-BERT), 0.8493 (KoELECTRA) and 0.8500 (KR-BERT-c), while the F1(macro) scores show a range from 0.6912 (Multi-BERT) to 0.7651 (KoBERT) with 4 epochs. We observe that all baselines achieve a similar performance, whereas Multi-BERT pre-trained on 104 languages present relatively lower performance. The KoELECTRA obtains overall the best or second best among six metrics, although this model has a seven times smaller parameter size (14M) than an average of other models (99M). This indicates the effects of the pre-training data source, considering that the KoELECTRA includes the corpus from Namuwiki and Modu that contain modern slang and buzzwords, while other models generally use Korean Wikipedia.

\paragraph{Evaluation for Multi-labels}
Table \ref{tab:breakdown_labels} shows the breakdown F1(micro) for multi-label classification from 1 to 4 labels\footnote{Further details are shown in Appendix Table \ref{tab:appendix_breakdown}.}. A single label task, achieving 0.8553 and 0.8490 from the KR-BERT-c and KoELECTRA, outperforms other multi-label tasks due to domain similarity. For the multi-label classification, KR-BERT-s achieved the best performance. It uses a sub-character tokenizer that can decompose Hangul(Korean language) syllable characters into sub-characters. Therefore, it provides greater granularity in detecting hate speech words, by identifying the sub-characters from different hate speech categories.\footnote{(e.g.) \begin{CJK}{UTF8}{mj}개빠ㄹ갱이년 = 개 ("dog" - \textit{profanity}) + 빠ㄹ갱이 ("communist" - \textit{politics}) + 년 ("bitch" - \textit{gender})\end{CJK}}.

\paragraph{Evaluation for Label-pairs}
Table \ref{tab:breakdown_pair} shows the F1-[macro, micro] scores for curated label pairs based on the proportion in the 2-labels classification. It illustrates that the KR-BERT-s model outperforms in six label pairs. In particular, it is very effective at detecting the \textit{origin and gender} pairs, achieving the highest F1 micro scores of 0.9494 across all label pairs and models. This model uses the sub-character-level tokenizer that can decompose various Korean characters (Hangul syllables) into sub-characters or graphemes to enable handling the \begin{CJK}{UTF8}{mj} bottom consonant (e.g. "gold-digger" [kko\#t\#baem] 꼬\#ㅊ\#뱀) or initial consonant (e.g. [k] ㅋ)\end{CJK}. This approach can detect new slang even if it is only a minor variation from other neutral words.


\section{Conclusion}
We propose K-MHaS, a new large-sized dataset for Korean hate speech detection with a multi-label annotation scheme. We provided extensive baseline experiment results, presenting the usability of a dataset to detect Korean language patterns in hate speech. In future work, the automatic hate speech moderation and counter-speech can be expanded. 



\section*{Acknowledgements}
We thank you the anonymous reviewers for their useful comments. This research is supported by an Australian Government Research Training Program (RTP) Scholarship.


\bibliography{anthology, custom}
\bibliographystyle{acl_natbib}

\appendix

\section{Appendix}
\label{sec:appendix}

\paragraph{Ethics/Broader Impact Statement}
The study follows the ethical policy set out in the ACL code of Ethics\footnote{https://www.aclweb.org/portal/content/acl-code-ethics} and addresses the ethical impact of presenting a new dataset. As described in the data section \ref{data_kmhas}, our annotated dataset is based on the online news comments data publicly available on Kaggle and Github. 
All annotators were recruited from a crowdsourcing platform. They were informed about hate speech before handling the data. Our instructions allowed them to feel free to leave if they were uncomfortable with the content. With respect to the potential risks, we note that the subjectivity of human annotation would impact on the quality of the dataset. 

\paragraph{The Korean language}
The Korean language is morphologically rich and the character structure is different to Latin-based language. A brief components used in the paper as follows:
\begin{itemize}
\begin{CJK}{UTF8}{mj}
\item \textbf{Consonant (자음)} : A consonant is a sound such as ‘p', ‘f', ‘n', or ‘t' which you pronounce by stopping the air flowing freely through your mouth.
\\- initial consonant (초성)
\\- bottom consonant (받침)
\item \textbf{Vowel (모음)} : A vowel is a sound such as the ones represented in writing by the letters ‘a', ‘e', ‘i', ‘o', and ‘u', which you pronounce with your mouth open, allowing the air to flow through it.
\item \textbf{Syllable (음절)} : A syllable is a part of a word that contains a single vowel sound and that is pronounced as a unit. So, for example, ‘book' has one syllable, and ‘reading' has two syllables.
\\- Korean romanization : (e.g. [kko\#t\#baem])
\\- Character level : (e.g. 꽃\#뱀)
\\- Sub-character level : (e.g. 꼬\#ㅊ\#뱀)
\end{CJK}
\end{itemize}

\paragraph{Implementation Details} 
For all baselines, we set the number of epochs as 4 and use a batch size of 32. For other hyper-parameters, we follow the configuration in the official GitHub implementation of the baselines. The source codes or pre-trained models for the baselines are available at the following GitHub addresses: Multilingual BERT\footnote{\begin{CJK}{UTF8}{mj}https://github.com/google-research/bert\end{CJK}}, KoELECTRA\footnote{\begin{CJK}{UTF8}{mj}https://github.com/monologg/KoELECTRA\end{CJK}}, KoBERT\footnote{\begin{CJK}{UTF8}{mj}https://github.com/SKTBrain/KoBERT\end{CJK}} and KR-BERT\footnote{\begin{CJK}{UTF8}{mj}https://github.com/snunlp/KR-BERT\end{CJK}}.



\paragraph{Experiments.}
A brief of tables displayed in Appendix as follows:
\begin{itemize}
\item Table \ref{tab:appendix_breakdown}: a breakdown of multi-label classification performance from 1 to 4 labels; 
\item Table \ref{tab:appendix_binary_overall}: overall binary classification performance;
\item Table \ref{tab:appendix_binary_breakdown}: a breakdown of binary classification performance;
\item Table \ref{tab:appendix_binary_breakdown}: F1 score for the top 10 three-label pairs in 3-labels classification;
\item Table \ref{tab:appendix_label_quadruplets}: F1 score for the top 5 four-label pairs in 4-labels classification.
\end{itemize}

\begin{table*}[t]
\centering
\resizebox{0.8\linewidth}{!}{
\renewcommand{\arraystretch}{1.1}
\begin{tabular}{clcccccc}
\hline
\textbf{\# Labels} & \textbf{Model} & \textbf{F1 (Macro)} & \textbf{F1 (Micro)} & \textbf{F1 (Weighted)} & \textbf{E.M.} & \textbf{AUC} & \textbf{H.L. (↓)} \\
\hline
\multirow{5}{*}{1} & BERT & 0.6666 & 0.8190 & 0.8202 & 0.7919 & 0.9011 & 0.0406 \\
 & KoELECTRA & 0.6953 & \underline{0.8490} & \underline{0.8508} & \textbf{0.8263} & \textbf{0.9213} & \underline{0.0341} \\
 & KoBERT & \underline{0.7321} & 0.8320 & 0.8370 & 0.8142 & 0.9110 & 0.0379 \\
 & KR-BERT(w. char) & \textbf{0.7336} & \textbf{0.8553} & \textbf{0.8543} & \underline{0.8239} & \underline{0.9145} & \textbf{0.0318} \\
 & KR-BERT(w. sub) & 0.6985 & 0.8392 & 0.8419 & 0.8062 & 0.9123 & 0.0360 \\
 \hline
\multirow{5}{*}{2} & BERT & 0.6389 & 0.8043 & 0.8174 & 0.5580 & 0.8524 & 0.0788 \\
 & KoELECTRA & \underline{0.6777} & 0.8612 & 0.8700 & 0.6511 & 0.8934 & 0.0577 \\
 & KoBERT & \textbf{0.7249} & \textbf{0.8854} & \textbf{0.8911} & \textbf{0.6794} & \textbf{0.9112} & \textbf{0.0482} \\
 & KR-BERT(w. char) & 0.6748 & 0.8405 & 0.8451 & 0.5912 & 0.8735 & 0.0642 \\
 & KR-BERT(w. sub) & 0.6718 & \underline{0.8703} & \underline{0.8723} & \underline{0.6535} & \underline{0.9000} & \underline{0.0542} \\
 \hline
\multirow{5}{*}{3} & BERT & 0.5784 & 0.7517 & 0.7522 & 0.2448 & 0.8040 & 0.1402 \\
 & KoELECTRA & 0.6146 & 0.7987 & 0.7953 & 0.3310 & 0.8362 & 0.1169 \\
 & KoBERT & \textbf{0.6523} & \underline{0.8290} & \underline{0.8251} & \textbf{0.3759} & \underline{0.8589} & \underline{0.1019} \\
 & KR-BERT(w. char) & 0.5828 & 0.7827 & 0.7732 & 0.2828 & 0.8239 & 0.1230 \\
 & KR-BERT(w. sub) & \underline{0.6164} & \textbf{0.8329} & \textbf{0.8263} & \underline{0.3586} & \textbf{0.8615} & \textbf{0.0996} \\
 \hline
\multirow{5}{*}{4} & BERT & 0.4776 & 0.7093 & 0.7029 & \textbf{0.1200} & 0.7610 & 0.2222 \\
 & KoELECTRA & 0.4511 & 0.7044 & 0.6639 & 0.0000 & 0.7680 & 0.2089 \\
 & KoBERT & 0.4177 & 0.6832 & 0.6460 & \underline{0.0400} & 0.7510 & 0.2267 \\
 & KR-BERT(w. char) & \underline{0.4837} & \underline{0.7439} & \underline{0.7226} & \textbf{0.1200} & \underline{0.7930} & \underline{0.1867} \\
 & KR-BERT(w. sub) & \textbf{0.5068} & \textbf{0.7771} & \textbf{0.7618} & \textbf{0.1200} & \textbf{0.8120} & \textbf{0.1733} \\
 \hline
\end{tabular}}
\caption{A breakdown of multi-label classification performance from 1 to 4 labels on K-MHaS for the five pre-trained language models at epoch 4 (E.M.:Exact Match, H.L.:Hamming Loss / KR-BERT (w. *): char = character-level, sub = sub-character-level)}
\label{tab:appendix_breakdown}
\end{table*}

\begin{table*}[t]
\centering
\resizebox{0.8\linewidth}{!}{
\renewcommand{\arraystretch}{1.1}
\begin{tabular}{l|cccccc}
\hline
\textbf{Model} & \textbf{F1 (Macro)} & \textbf{F1 (Micro)} & \textbf{F1 (Weighted)} & \textbf{E.M.} & \textbf{AUC} & \textbf{H.L. (↓)} \\ \hline
BERT & 0.8495 & 0.8507 & 0.8505 & 0.8507 & 0.8488 & 0.1493 \\
KoELECTRA & 0.8756 & 0.8766 & 0.8765 & 0.8766 & 0.8750 & 0.1234 \\
KoBERT & 0.8687 & 0.8692 & 0.8693 & 0.8692 & 0.8696 & 0.1308 \\
KR-BERT (w. char) & \underline{0.8846} & \underline{0.8850} & \underline{0.8851} & \underline{0.8850} & \textbf{0.8862} & \underline{0.1150} \\
KR-BERT (w. sub) & \textbf{0.8869} & \textbf{0.8879} & \textbf{0.8877} & \textbf{0.8879} & \underline{0.8857} & \textbf{0.1121} \\ \hline
\end{tabular}}
\caption{Overall binary classification performance on the K-MHaS dataset for the five pre-trained language models at epoch 4 (E.M.:Exact Match, H.L.:Hamming Loss / KR-BERT (w. *): char = character-level, sub = sub-character-level)}
\label{tab:appendix_binary_overall}
\end{table*}

\begin{table*}[t]
\centering


\resizebox{0.8\linewidth}{!}{
\renewcommand{\arraystretch}{1.1}
\begin{tabular}{c|l|ccccc}
\hline
\textbf{Label} & \textbf{Model} & \textbf{F1 (Macro)} & \textbf{F1 (Micro)} & \textbf{F1 (Weighted)} & \textbf{E.M.} & \textbf{H.L. (↓)} \\
\hline
\multirow{5}{*}{Hate Speech} & BER & 0.4518 & 0.8243 & 0.9037T & 0.8243 & 0.1757 \\
 & KoELECTRA & 0.4606 & 0.8540 & \underline{0.9212} & 0.8540 & 0.1460 \\
 & KoBERT & \underline{0.4666} & \underline{0.8746} & \textbf{0.9331} & \underline{0.8746} & \underline{0.1254} \\
 & KR-BERT (w. char) & 0.4611 & 0.8558 & 0.7892 & 0.8558 & 0.1442 \\
 & KR-BERT (w. sub) & \textbf{0.4724} & \textbf{0.8953} & 0.8458 & \textbf{0.8953} & \textbf{0.1047} \\ \hline
\multirow{5}{*}{None} & BERT & 0.4662 & 0.8733 & \underline{0.9323} & 0.8733 & 0.1267 \\
 & KoELECTRA & \underline{0.4726} & \underline{0.8960} & \textbf{0.9452} & \textbf{0.8960} & \underline{0.1040} \\
 & KoBERT & 0.4637 & 0.8645 & 0.9273 & 0.8645 & 0.1355 \\
 & KR-BERT (w. char) & \textbf{0.4772} & \textbf{0.9126} & 0.8709 & \textbf{0.9126} & \textbf{0.0874} \\
 & KR-BERT (w. sub) & 0.4687 & 0.8821 & 0.8268 & 0.8821 & 0.1179 \\ \hline
\end{tabular}}
\caption{A breakdown of binary classification performance on the K-MHaS dataset for the five pre-trained language models at epoch 4 (E.M.:Exact Match, H.L.:Hamming Loss / KR-BERT (w. *): char = character-level, sub = sub-character-level, bi = Bidirectional WordPiece tokenizer)}
\label{tab:appendix_binary_breakdown}
\end{table*}

\begin{table*}[t]
\centering
\resizebox{\textwidth}{!}{
\renewcommand{\arraystretch}{1.2}
\begin{tabular}{l|c|ccccc|ccccc}
\hline
\textbf{Label Triplets} & \textbf{\# triplets} & \multicolumn{5}{c|}{\textbf{F1 (macro)}} & \multicolumn{5}{c}{\textbf{F1 (micro)}} \\
\hline
\multicolumn{2}{l|}{\multirow{2}{*}{\textbf{Overall Performance (F1)}}} & \textbf{BERT} & \textbf{KoELECTRA} & \textbf{KoBERT} & \textbf{KR-BERT-c} & \textbf{KR-BERT-s} & \textbf{BERT} & \textbf{KoELECTRA} & \textbf{KoBERT} & \textbf{KR-BERT-c} & \textbf{KR-BERT-s} \\ \cline{3-12}
\multicolumn{2}{l|}{} & 0.6912 & 0.7245 & \textbf{0.7651} & \underline{ 0.7444} & 0.7245 & 0.8139 & \underline{ 0.8493} & 0.8413 & \textbf{0.8500} & 0.8445 \\
\hline
\textit{Origin \& Politics \& Profanity} & 41 & 0.2780 & 0.2935 & \underline{ 0.2954} & 0.2937 & \textbf{0.3125} & 0.8224 & 0.8739 & \underline{ 0.8869} & 0.8739 & \textbf{0.9316} \\
\textit{Politics \& Profanity \& Age} & 37 & 0.2781 & \underline{ 0.3054} & \textbf{0.3174} & 0.2981 & 0.2971 & 0.8205 & \underline{ 0.9126} & \textbf{0.9395} & 0.8867 & 0.8824 \\
\textit{Physical \& Politics \& Profanity} & 32 & 0.2483 & 0.2809 & \textbf{0.2960} & 0.2823 & \underline{ 0.2939} & 0.7296 & 0.8304 & \underline{ 0.8750} & 0.8421 & \textbf{0.8764} \\
\textit{Origin \& Profanity \& Gender} & 30 & \underline{ 0.2545} & 0.2314 & 0.2527 & 0.2463 & \textbf{0.2886} & 0.7467 & 0.7397 & 0.7368 & \underline{ 0.7671} & \textbf{0.8712} \\
\textit{Physical \& Profanity \& Gender} & 24 & 0.2151 & 0.2665 & \underline{ 0.2730} & 0.2459 & \textbf{0.2811} & 0.6306 & 0.7869 & \underline{ 0.8226} & 0.7521 & \textbf{0.8413} \\
\textit{Origin \& Physical \& Profanity} & 14 & 0.2692 & 0.2811 & \textbf{0.2873} & 0.2351 & \underline{ 0.2865} & 0.7532 & \underline{ 0.8378} & 0.8312 & 0.7273 & \textbf{0.8421} \\
\textit{Politics \& Age \& Gender} & 14 & 0.2593 & \textbf{0.2933} & \underline{ 0.2830} & 0.2319 & 0.2406 & 0.7606 & \textbf{0.8684} & \underline{ 0.8158} & 0.6970 & 0.7059 \\
\textit{Profanity \& Age \& Gender} & 13 & 0.2686 & 0.2712 & \underline{ 0.2932} & 0.2593 & \textbf{0.2695} & 0.8182 & 0.8060 & \textbf{0.8732} & 0.8000 & \underline{ 0.8358} \\
\textit{Origin \& Physical \& Gender} & 12 & \underline{ 0.2692} & 0.2327 & 0.2407 & 0.2347 & \textbf{0.2703} & \underline{ 0.7812} & 0.6780 & 0.7458 & 0.7143 & \textbf{0.8065} \\
\textit{Origin \& Age \& Gender} & 10 & 0.2736 & \underline{ 0.2781} & \textbf{0.3093} & 0.2428 & 0.2411 & \underline{ 0.8235} & 0.8302 & \textbf{0.9123} & 0.7347 & 0.7600 \\ \hline
\end{tabular}}
\caption{F1 score for the top 10 three-label pairs on the K-MHaS dataset for the five pre-trained language models at epoch 4 (\# total label triplets = 290 / KR-BERT-*: c = character-level WordPiece tokenizer, s = sub-character-level WordPiece tokenizer)}
\label{tab:appendix_label_triplets}
\end{table*}

\begin{table*}[t]
\centering
\resizebox{\textwidth}{!}{
\renewcommand{\arraystretch}{1.2}
\begin{tabular}{ll|c|ccccc|ccccc}
\hline
\multicolumn{2}{l|}{\textbf{Label Quadruplets}} & \textbf{\# quadruplets} & \multicolumn{5}{c}{\textbf{F1 (macro)}} & \multicolumn{5}{c}{\textbf{F1 (micro)}} \\
\hline
\multicolumn{3}{l|}{\multirow{2}{*}{\textbf{Overall Performance (F1)}}} & \multicolumn{1}{c}{\textbf{BERT}} & \multicolumn{1}{c}{\textbf{KoELECTRA}} & \multicolumn{1}{c}{\textbf{KoBERT}} & \multicolumn{1}{c}{\textbf{KR-BERT-c}} & \multicolumn{1}{c}{\textbf{KR-BERT-s}} & \multicolumn{1}{c}{\textbf{BERT}} & \multicolumn{1}{c}{\textbf{KoELECTRA}} & \multicolumn{1}{c}{\textbf{KoBERT}} & \multicolumn{1}{c}{\textbf{KR-BERT-c}} & \multicolumn{1}{c}{\textbf{KR-BERT-s}} \\ \cline{4-13}
\multicolumn{3}{l|}{} & \multicolumn{1}{c}{0.4776} & \multicolumn{1}{c}{0.4511} & \multicolumn{1}{c}{0.4177} & \underline{0.4837} & \textbf{0.5068} & \multicolumn{1}{c}{0.7093} & \multicolumn{1}{c}{0.7044} & \multicolumn{1}{c}{0.6832} & \underline{0.7439} & \textbf{0.7771} \\ \hline
\multicolumn{2}{l|}{\textit{Origin \& Profanity \& Age \& Gender}} & 5 & \textbf{0.3567} & 0.2950 & 0.2346 & 0.2932 & \underline{0.3086} & \textbf{0.8000} & 0.6875 & 0.5806 & 0.7500 & \underline{0.7879} \\
\multicolumn{2}{l|}{\textit{Origin \& Physical \& Profanity \& Gender}} & 4 & 0.2582 & 0.2804 & 0.2508 & \underline{0.3302} & \textbf{0.3757} & 0.5385 & 0.7200 & 0.6400 & \underline{0.7692} & \textbf{0.8571} \\
\multicolumn{2}{l|}{\textit{Origin \& Physical \& Politics \& Profanity}} & 3 & \textbf{0.4000} & \underline{0.3333} & 0.3111 & 0.3333 & \underline{0.3667} & \textbf{0.9091} & 0.8000 & 0.7368 & 0.8000 & \underline{0.8571} \\
\multicolumn{2}{l|}{\textit{Origin \& Politics \& Profanity \& Age}} & 3 & \underline{0.4222} & 0.3444 & 0.3667 & \textbf{0.4444} & \textbf{0.4444} & 0.8800 & 0.8000 & 0.8571 & \textbf{1.0000} & \underline{0.9600} \\
\multicolumn{2}{l|}{\textit{Origin \& Politics \& Profanity \& Gender}} & 2 & 0.1852 & \textbf{0.2963} & \underline{0.2593} & \textbf{0.2963} & \textbf{0.2963} & 0.5455 & \textbf{0.7692} & \underline{0.6667} & \underline{0.6667} & \textbf{0.7692} \\ \hline
\end{tabular}}
\caption{F1 score for the top 5 four-label pairs on the K-MHaS dataset for the five pre-trained language models at epoch 4 (\# total label quadruplets = 25 / KR-BERT-*: c = character-level WordPiece tokenizer, s = sub-character-level WordPiece tokenizer)}
\label{tab:appendix_label_quadruplets}
\end{table*}

\end{document}